\documentclass[11pt]{article}

\usepackage[preprint]{acl}

\usepackage{times}
\usepackage{latexsym}

\usepackage[T1]{fontenc}

\usepackage[utf8]{inputenc}

\usepackage{microtype}

\usepackage{inconsolata}

\usepackage{graphicx}
\graphicspath{{../figs/}}

\usepackage{amsmath,amssymb}
\usepackage{booktabs}
\usepackage{multirow}
\usepackage{array}
\usepackage{enumitem}
\usepackage[most]{tcolorbox}

\newcommand{\Df}{\mathcal{D}_f}

\newcommand{\PPL}{\mathrm{PPL}}
\newcommand{\Lone}{L_1}
\newcommand{\Ltwo}{L_2}
\newcommand{\Lthree}{L_3}

\usepackage[table]{xcolor}
\usepackage{subcaption}
\definecolor{bestcell}{HTML}{E8F4EA}

\title{{\sc PreUnlearn}: Auditing Collateral Knowledge Damage \\Before Large Language Model Unlearning}

\author{Bo Su \\
Indiana University \\
Bloomington, IN, USA \\
\texttt{subo@iu.edu}
\And
Ankit Shah \\
Indiana University \\
Bloomington, IN, USA \\
\texttt{ankit@iu.edu}
\And
Thai Le \\
Indiana University \\
Bloomington, IN, USA \\
\texttt{tle@iu.edu}}

\begin{document}
\maketitle
\begin{abstract}

Machine unlearning for large language models (LLMs) aims to remove specified knowledge while preserving the rest of the model's capabilities. However, the boundary between knowledge to forget and knowledge to retain is often unclear, since related and even distant information may be entangled in the model. In this paper, we study LLM unlearning from a data-centric perspective and measure how unlearning effects propagate from the forget set to same-domain and distant-domain knowledge not after but before unlearning. We find a consistent decay pattern: collateral damage is strongest near the forget set, weakens with semantic distance, but does not disappear at domain boundaries. We further ask whether such damage can be audited before unlearning is executed. We formulate forget-set auditing as a pre-unlearning prediction task and analyze which data features are most predictive of downstream damage. Our results show that interaction features between the forget set and evaluation set provide the strongest signals, suggesting that collateral damage is partly reflected in data geometry before model updates occur. These findings position forget-set auditing as an early warning tool for identifying risky unlearning runs and designing more reliable unlearning procedures. Code and data are available at \href{https://github.com/BartSu/PreUnlearn}{https://github.com/BartSu/PreUnlearn}.

\end{abstract}

\section{Introduction}
\label{sec:intro}

\begin{figure}[tb!]
\centering
\makebox[\columnwidth][c]{\includegraphics[width=1.2\columnwidth]{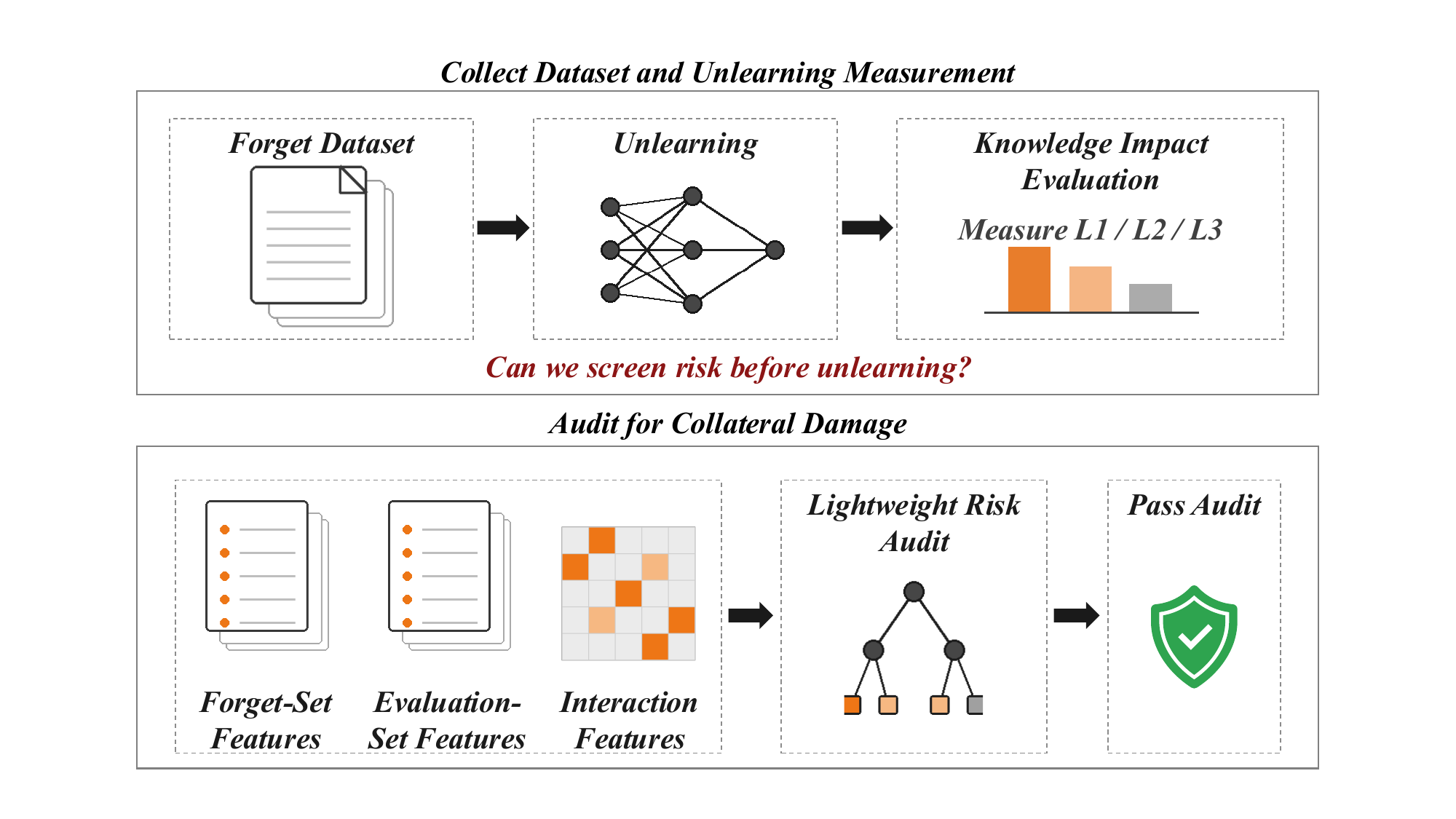}}
\caption{PreUnlearn framework with two stages: \textbf{(1) Three-layer Unlearning Impact} where an LLM is unlearned on a candidate forget set $\Df$ and scored across three semantic layers ($\Lone, \Ltwo,\Lthree$) of decreasing relevancy to $\Df$ and ultimately, \textbf{(2) Pre-unlearn Impact Auditing} produces a lightweight auditor to estimate per-layer potential collateral damage risk on retain knowledge, screening the candidate $\Df$ \emph{before} unlearning.}
\label{fig:hero}
\end{figure}

As large language models (LLMs) are increasingly deployed in high-stakes settings, machine unlearning has become essential for removing sensitive, harmful, outdated, or legally restricted information while preserving overall model utility. Practical motivations for unlearning span privacy regulations, copyright disputes, user deletion requests, and the removal of toxic or unsafe content~\citep{llm-unlearning-survey, machine-unlearning, maini2024tofu, shi2024muse, li2024wmdp, dorna2025openunlearning}. In some applications, unlearning may also be required to suppress domain-specific capabilities associated with safety risks, including offensive cybersecurity knowledge such as exploit-development procedures, vulnerability analysis, or other misuse-prone behaviors. Since retraining from scratch is typically impractical, LLM unlearning seeks to suppress target knowledge through post-training updates \citep{whp, jang2023ga, li2024wmdp} while preserving the model's remaining capabilities and factual knowledge. This preservation requirement is central to practical deployment: an unlearning procedure is incomplete if the same update unintentionally degrades neighboring facts, distant knowledge, or otherwise unrelated capabilities. Hence, it is critical to evaluate whether unlearning targeted knowledge (the forget set) inadvertently affects semantically related or unrelated knowledge elsewhere in the model, which is the central focus of this study (see Figure~\ref{fig:hero}). 

Existing unlearning evaluations do not yet characterize collateral damage with sufficient
granularity. Most prior studies report aggregate utility metrics or a limited set of post-hoc
probes, which often fail to capture the data-centric structure of unlearning damage, including how performance degradation propagates as evaluation data moves from the forget set itself, to same-domain knowledge, and further toward distantly related or even orthogonal knowledge. Recent work has shown that standard utility benchmarks can remain deceptively high even when same-domain or distant-domain knowledge has already been substantially corrupted \citep{ko2025probing}, raising concerns about the adequacy of current evaluation practices. At the same time, most existing benchmarks assume a fixed forget set and implicitly treat the choice of evaluation set as given. Consequently, two practical research questions (RQs) remain largely unexplored: \textit{``How will unlearning with forget set X affect the model on knowledge Y?'' (\ref{rq1})}, and then  \textit{``Can we predict \textit{ahead of time} such a potential impact or collateral damage even before unlearning?'' (\ref{rq2})} This will enable practitioners to anticipate high-risk unlearning runs before expensive optimization is performed.

\begin{tcolorbox}[
    colback=gray!10, colframe=gray!50,
    arc=2mm, boxrule=0.4pt,
    left=6pt, right=6pt, top=4pt, bottom=4pt,
]

\begin{enumerate}[leftmargin=\dimexpr\parindent+1.8\labelwidth\relax,
itemsep=2pt,topsep=0pt,label=\textbf{RQ.\arabic*.}]
    \item \textbf{(Three-layer Unlearning Impact)} - How does unlearning impact spread from the forget set X to same-domain and distant-domain knowledge Y? \label{rq1}
    \item \textbf{(Pre-unlearn Impact Auditing)} - Can we predict the collateral damage would result knowledge Y from unlearning knowledge X even before unlearning?\label{rq2}
\end{enumerate}
\end{tcolorbox}

We address these two questions by formulating pre-unlearning auditing as a supervised modeling problem (Fig.~\ref{fig:hero}). Organizing a dataset into $\Lone$ (intended degradation), $\Ltwo$ (same-domain damage), and $\Lthree$ (irrelevant-domain damage), we predict future unlearning impact using features of the forget set, the evaluation set, their interaction. This formulation uses prediction not as an end in itself, but as a tool to identify which pre-unlearning signals explain later collateral damage.

The measurements reveal a consistent but imperfect decay pattern: unlearning impact is strongest on forget set, weaker on same-domain knowledge, and weakest, but still present, on distant-domain knowledge. The audit further shows that interaction features between the forget and evaluation sets, such as semantic proximity, representation-shape ratios, and lexical or length relationships, are especially predictive and remain stable across unlearning algorithms.

\noindent\textbf{Our main contributions are:}
\begin{enumerate}[leftmargin=*,itemsep=2pt,topsep=2pt]
    \item \textbf{Three-layer measurement framework.} We organize unlearning impact into intended ($L_1$), same-domain ($L_2$), and distant-domain ($L_3$) degradation, and show across two model families and three algorithms that damage consistently decays with semantic distance but remains visible beyond the direct target, with substantial variation across forget sets under fixed hyperparameters.
    \item \textbf{Pre-unlearning auditing as supervised prediction.} We formulate forget-set auditing as a regression problem over (forget, evaluation) pairs, using only pre-update features of the data, with no access to gradients, unlearned checkpoints, or post-hoc measurements.
    \item \textbf{Empirical characterization of predictive signals.} We show that cross-set geometric features (centroid distance, similarity, length and lexical ratios) dominate over intrinsic properties of either set, remain stable across unlearning algorithms, and yield ranking quality strong enough for practical triage.
\end{enumerate}

\section{Related Work}
\label{sec:related}
\subsection{LLM Unlearning}
LLM unlearning aims to remove selected knowledge from a pretrained
model while preserving overall utility \cite{llm-unlearning-survey}.
Existing work follows two paradigms: \textit{fine-tuning-then-unlearning},
where the forget set is a subset of a known fine-tuning corpus
(e.g., TOFU, MUSE, FIUBench \cite{maini2024tofu, shi2024muse, FIUBench, he2026order}),
and \textit{direct unlearning}, where the target knowledge is already
embedded in the pretrained model (e.g., WMDP, RWKU \cite{li2024wmdp,
jin2024rwku}). Our setting follows the latter, which is closer to
real deployment.


\begin{figure*}[tb]
\centering
\footnotesize

\begin{tikzpicture}[
    x=1cm,y=1cm,
    every node/.style={font=\footnotesize},
    box/.style={
        draw,
        rounded corners=2pt,
        thick,
        align=center,
        minimum width=1.9cm,
        minimum height=0.85cm,
        inner sep=2pt
    },
    arrow/.style={thick,->,>=stealth}
]

\node[box, fill=blue!8]   (raw)     at (0,0)    {WikiText-103\\raw passages};
\node[box, fill=cyan!10]  (filter)  at (2.8,0)  {Filter usable\\passages};
\node[box, fill=teal!10]  (embed)   at (5.6,0)  {Embed\\passages};
\node[box, fill=green!10] (cluster) at (8.4,0)  {Cluster semantic\\passage pools};
\node[box, fill=orange!12](noise)   at (11.2,0) {Exclude \\ noise };
\node[box, fill=purple!10](sample)  at (14.0,0) {Sample unlearn\\datasets};

\draw[arrow] (raw) -- (filter);
\draw[arrow] (filter) -- (embed);
\draw[arrow] (embed) -- (cluster);
\draw[arrow] (cluster) -- (noise);
\draw[arrow] (noise) -- (sample);

\end{tikzpicture}

\caption{Dataset construction schema. WikiText-103 passages are filtered, embedded, clustered into semantic passage pools, and then sampled into unlearning datasets. Each dataset contains disjoint \emph{forget} and \emph{retain} splits, which later support direct unlearning and three-layer impact evaluation.}
\label{fig:dataset-pipeline}
\end{figure*}
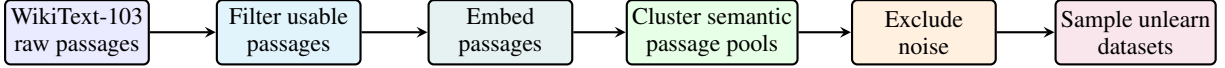

\subsection{Collateral Damage in LLM Unlearning}
Unlearning can degrade knowledge beyond the forget set $\Df$.
\citet{ko2025probing} introduce knowledge hole probing and show
that static benchmarks such as MMLU \cite{mmlu} and TruthfulQA
\cite{truthfulqa} can miss collateral damage incurred by unlearning,
motivating evaluations that go beyond whether $\Df$ is removed.

A related question is which forget sets are likely to cause such
damage. Prior work studies which data is hardest to remove or
induces the largest side effects \citep{thudi2022unrolling,
kurmanji2024towards}. Influence-based selection \citep{koh2017understanding}
is related but requires model gradients, scales poorly to LLMs,
and yields per-example rather than forget-set-level scores. Closer
to our approach, a data-centric line predicts downstream behavior
from dataset properties \citep{dang2024curious, ilyas2022datamodels, he2026order};
we extend this to pre-unlearning auditing over forget--evaluation pairs.

\section{\ref{rq1} Three-layer Unlearning Impact}

\subsection{Problem Formulation}
\label{sec:problem}

Let $M_{\theta_0}$ denote the target LLM before unlearning, and let $\mathcal{U}(\cdot)$ denote a fixed post-training unlearning algorithm. We consider a collection of candidate semantic domains $\mathcal{G}=\{G_i\}_{i=1}^{N}$, where each domain contains documents covering a coherent semantic topic. For each domain $G_i$, we construct a forget subset $G_i^{\mathrm{f}} \subset G_i$, which serves as the candidate forget set for unlearning. For each candidate domain $G_i$, we define the forget set as $D_f \leftarrow G_i^{\mathrm{f}}$ and obtain the corresponding unlearned checkpoint:
\[
M^*_{\theta}
\leftarrow
\mathcal{U}(M_{\theta}, D_f).
\]
Throughout this work, we fix the base model $M_{\theta_0}$ and $\mathcal{U}$, while varying only the semantic content of the forget set $D_f$. This allows us to isolate and analyze how unlearning different semantic domains affects the resulting model behavior and knowledge impact patterns.

For each unlearning run, we compare the unlearned model $M^*_{\theta}$ against the original model $M_{\theta}$ on a shared evaluation set constructed from all domains $G_i$. This shared
evaluation design then lets us ask where the impact of unlearning $D_f$ leaves on a three-layer knowledge impact to be described below.

\paragraph{Three-layer Knowledge Impact.} To characterize the effect of unlearning, we organize post-unlearning knowledge degradation into three semantic layers. Given a forget set $D_f$, the first layer measures degradation on the forget set itself, denoted as $\Lone$, corresponding to the intended effect of unlearning. The second layer measures degradation on held-out passages that are semantically close to the forget domain, denoted as $\Ltwo$, capturing local collateral damage on related knowledge. The third layer measures degradation on passages drawn from other semantic domains, denoted as $\Lthree$, capturing unintended forgetting on distant and irrelevant knowledge. We can then summarize the resulting three-layer collateral profile as:
\[
\mathbf{y} = (\Lone, \Ltwo, \Lthree)
\]

\subsection{Experimental Setup}
\label{sec:setup}

\paragraph{Dataset Preparation.}
\label{sec:dataset}
We use WikiText-103~\citep{merity2017pointer}, which was constructed from Wikipedia
articles, for unlearning because Wikipedia text is
widely used in LLM pretraining, making it reasonable to assume that the target
models have already learned much of this content. We confirmed this
assumption by observing consistently low PPL on sampled WikiText-103 passages.
Thus, WikiText-103 provides a suitable testbed for studying unlearning on
knowledge that is plausibly already present in the models.

We process WikiText-103 via a quality control pipeline (Fig. ~\ref{fig:dataset-pipeline}), resulting in $10$ \textit{well-separated} semantic clusters used to construct forget-set candidates. From the yielding clusters, we construct $100$ unlearning datasets, $10$ for each cluster, each of which contains two same-cluster but mutually disjoint splits of $50$ texts: a forget set $D_f$, which is the direct unlearning target, and a retain set $D_r$, which
provides same-domain text that should remain usable after unlearning. This forget/retain construction matches the standard unlearning setup: unlearning is to remove $D_f$ while preserving performance on $D_r$. Details on the dataset construction are provided in the Appendix \ref{app:repro}.

\begin{figure*}[t]
  \centering
  \includegraphics[width=\textwidth]{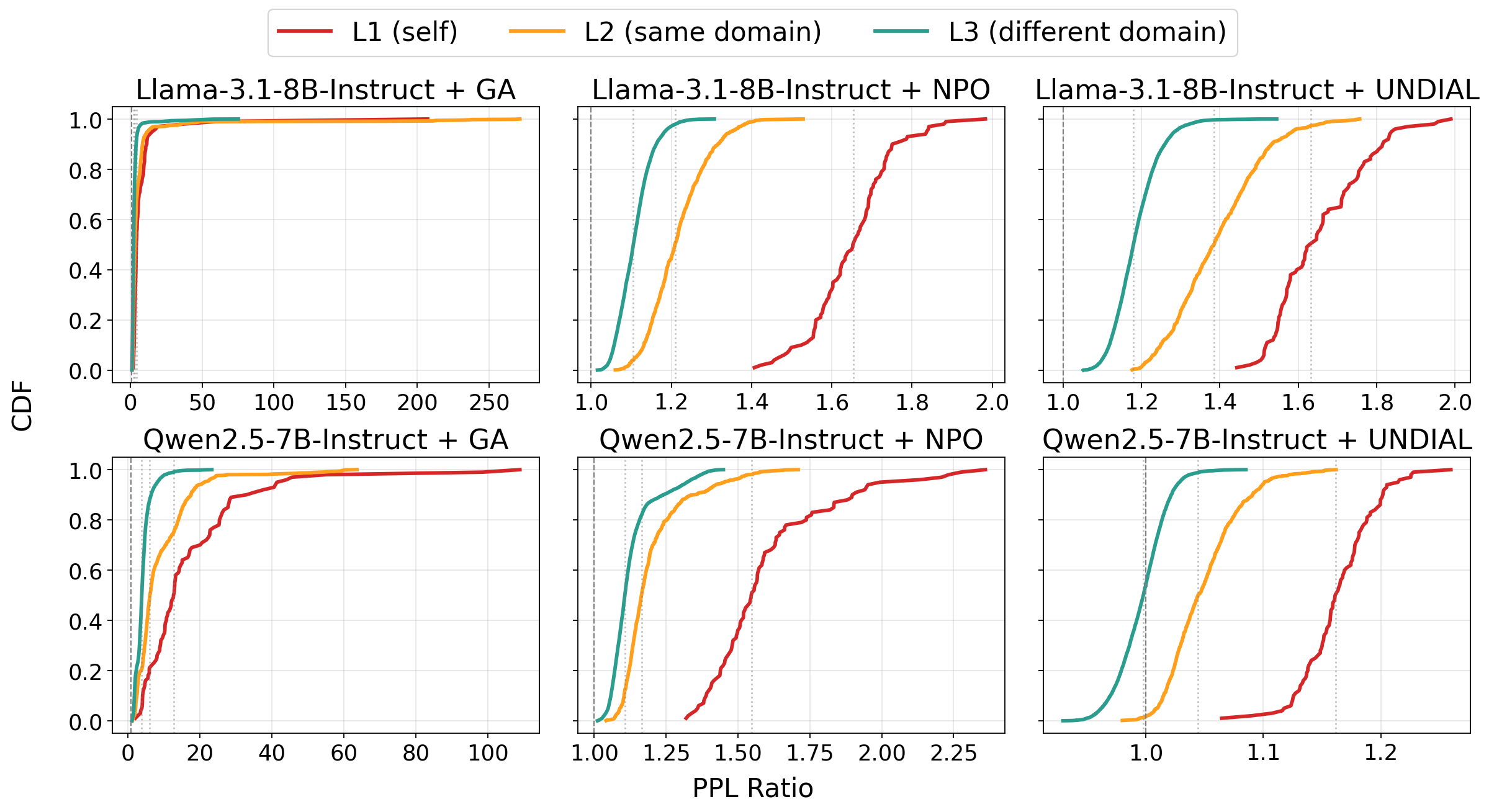}
  \caption{\textbf{PPL ratio distribution by layer across six settings.}
  CDF of PPL ratio (after / before unlearning) for $\Lone$ (self),
  $\Ltwo$ (same domain), $\Lthree$ (different domain). Dotted grey verticals
  mark per-layer medians; dashed line at ratio $=1$ is the no-change reference.}
  \label{fig:layered-all}
  \end{figure*}

Let $\mathcal{I}(\cdot)$ denote the unlearning impact metric, which we will later define, we can define the three evaluation layers $\Lone$, $\Ltwo$, and $\Lthree$ as follows:
\[
\Lone{=}\mathcal{I}(D_f), \;
\Ltwo{=}\mathcal{I}(G_i \setminus D_f), \;
\Lthree{=}\mathcal{I}\Big(\bigcup_{j \neq i} G_j\Big)
\]

\paragraph{Target Unlearning Model.}
We use two open-weight instruction-tuned target models:
Llama-3.1-8B-Instruct \citep{dubey2024llama3} and Qwen2.5-7B-Instruct
\citep{qwen2025qwen25}. As a sanity check, we measured base-model PPL on sampled WikiText-103 passages and observed low values, which provides indirect evidence that these passages are familiar to the target models. Using two model families lets us check whether the results are specific to a single architecture family.

\paragraph{Unlearning Algorithm.} We evaluate three unlearning algorithms Gradient Ascent (GA)~\citep{jang2023ga}, Negative Preference Optimization (NPO)\citep{zhang2024npo} and Unlearning via Self-Distillation on Adjusted Logits (UNDIAL)\citep{dong2025undial} because they can represent three distinct families of LLM unlearning methods.

\paragraph{Evaluation and Metrics.}
For an evaluation set $S$, we measure the impact of unlearning $D_f$ via $U$ using the perplexity ratio:
\begin{align}
    R(S)
    =
    \frac{\PPL(M^*_{\theta}, x)}
         {\PPL(M_{\theta}, x)} \geq 1.0,
\end{align}
where the degree by which $R(S)$ increases from $1.0$ indicates the impact of unlearning or how much the model forgets on average the knowledge of each passage $x{\in}S$. 

\begin{figure*}[tb!]
\centering
\begin{subfigure}[t]{0.495\linewidth}
    \centering
    \includegraphics[width=\linewidth]{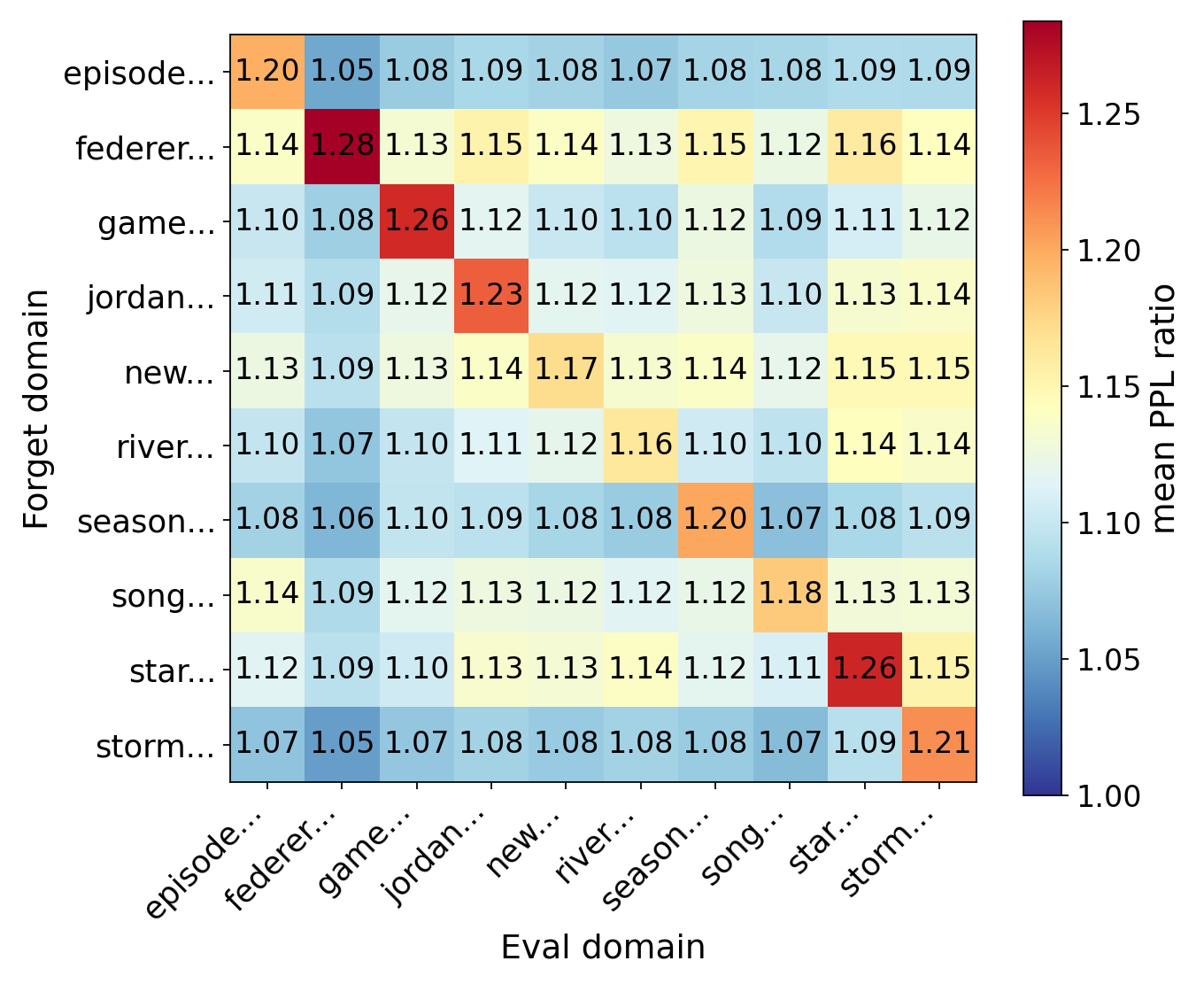}
    \caption{NPO}
    \label{fig:npo_llama_heatmap}
\end{subfigure}
\hfill
\begin{subfigure}[t]{0.495\linewidth}
    \centering
    \includegraphics[width=\linewidth]{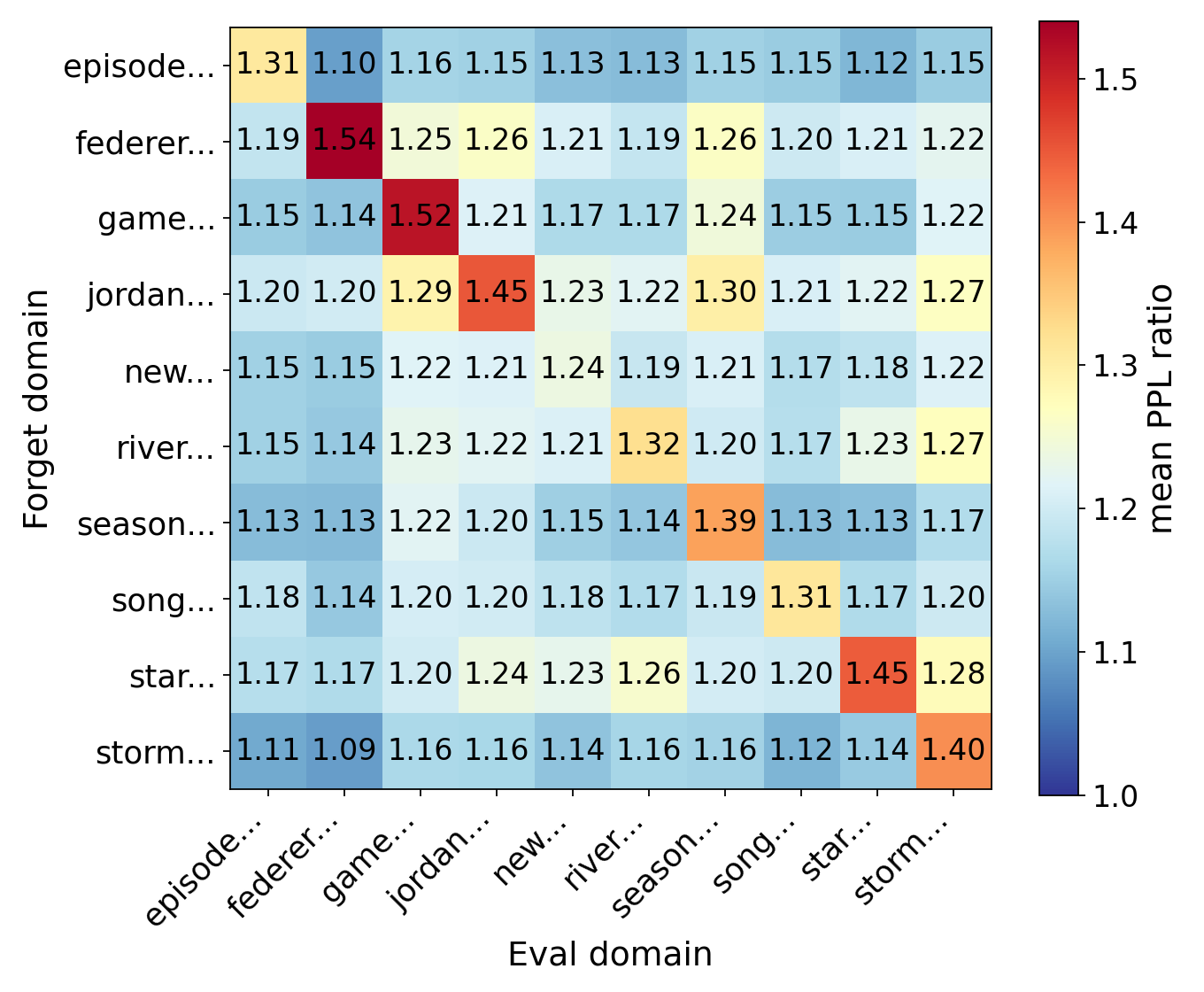}
    \caption{UNDIAL}
    \label{fig:undial_llama_heatmap}
\end{subfigure}
\caption{Mean PPL ratio per domain pair on Llama-3.1-8B-Instruct under NPO and UNDIAL.}
\label{fig:llama_domain_heatmaps}
\end{figure*}

\subsection{Experiment Results}
\label{sec:three-layer}

\setlength{\tabcolsep}{1pt}

Figure~\ref{fig:layered-all} reports the profile across all model$\times$algorithm
settings, from which we draw three distinct observations and findings:

\textbf{{Finding \#1: Unlearning impact decays with distance from the forget set.}}
For every model$\times$algorithm pairs, the average
PPL ratio satisfies $\Lone{>}\Ltwo{>}\Lthree$. For instance,
Llama-3.1-8B with GA yields
$\Lone{=}5.20{\times}{>}\Ltwo{=}3.80{\times}{>}\Lthree{=}2.42{\times}$, an
across-layer spread of $2.15{\times}$ (Figure~\ref{fig:layered-all}). This monotonic decay holds in all settings, indicating that unlearning induces geometry-dependent ripple effects that spread outward from the forget set.

\textbf{{Finding \#2: Impact magnitude is algorithm-dependent.}} GA produces the most aggressive impact, reaching $\Lone{=}12.85\times$ on Qwen2.5-7B, whereas NPO, UNDIAL remain within $1.05$--$1.65\times$ at $\Lone$ across base models. Consistent with GA's lack of a regularizer on a retain set, this gap shows that the choice of algorithm is determinant of impact magnitude, while the layered ordering of {Finding \#1} is preserved across the full range.

\textbf{{Finding \#3: The impact on three-layer profiles varies with the forget set.}} The same algorithm and hyper-parameters do not produce a single fixed impact profile. Across the $100$ forget sets, the spread is $2.15{\times}$ (max/min) under the Llama-3.1-8B when unlearning with GA
(Figure~\ref{fig:layered-all}). This variation makes the forget set itself worth auditing before paying for full unlearning and cross-evaluation. 

\textbf{{Finding \#4: Layer-wise impact separates at the distribution level.}} The empirical CDFs in Fig.~\ref{fig:layered-all} show a clear
distribution-level ordering: $L_1$ is shifted furthest to the right,
followed by $L_2$ and then $L_3$, across nearly all six settings.
This separation persists across quantiles, showing that the layered
structure is not only an average-level phenomenon. While a small fraction
of points, especially in the farther layers, remain near or slightly below
the no-change reference ratio $1$, the distributions also exhibit
substantial right tails. In particular, the GA settings contain many
evaluation points with very large PPL-ratio increases, indicating that
unlearning can induce severe collateral damage rather than merely mild
average degradation. 

\textbf{{Finding \#5: The layered unlearning pattern persists with domain-pair heterogeneity.}}
Figure~\ref{fig:layered-all} shows that $\Lone{>}\Ltwo{>}\Lthree$ holds on average, and
Figure~\ref{fig:llama_domain_heatmaps} shows that this pattern persists across domain pairs. We aggregate
the $100{\times}100$ evaluation matrix into $10{\times}10$ heatmaps for NPO and UNDIAL on Llama-3.1-8B-
Instruct, where diagonal cells denote matched forget--evaluation domains and off-diagonal cells denote
cross-domain evaluation. In both heatmaps, diagonal cells are consistently stronger than off-diagonal cells,
showing that same-domain knowledge is more affected than distant-domain knowledge. The heatmaps also reveal
domain-pair heterogeneity: some forget domains induce broader off-domain damage, and some evaluation domains
are more vulnerable.

\subsection{Behavioral Validation Beyond Perplexity}
\label{sec:qa_probe}

Perplexity provides a scalable token-level measure of degradation, but
it does not by itself establish whether the observed changes correspond
to a loss of answerable knowledge. We therefore conduct a focused
behavioral validation using a closed-book multiple-choice QA probe
synthesized from the corresponding WikiText passages. This experiment
uses Llama-3.1-8B-Instruct with GA and evaluates the same already-trained
unlearned checkpoints used in the perplexity analysis.

As shown in Table~\ref{tab:qa_probe}, the accuracy decrease is largest
for L1, followed by L2 and L3, matching the ordering of the corresponding
PPL ratios. Specifically, accuracy decreases by 9.3 percentage points on
L1, 3.6 points on L2, and 2.9 points on L3. This result provides a
targeted behavioral check that the observed perplexity degradation is
accompanied by reduced answer correctness in this setting. Because the
probe covers one model-algorithm configuration, we do not interpret it
as a comprehensive validation of perplexity across all unlearning
settings.

\begin{table}[t]
\centering
\small
\begin{tabular}{@{}lrrrr@{}}
\toprule
Layer & PPL ratio & Base acc (\%) & Unlearned acc (\%) & Drop(\%) \\
\midrule
L1 & $4.97\times$ & 61.4 & 52.1 & $-9.3$ \\
L2 & $3.49\times$ & 57.9 & 54.3 & $-3.6$ \\
L3 & $2.31\times$ & 59.0 & 56.2 & $-2.9$ \\
\bottomrule
\end{tabular}
\caption{
Closed-book multiple-choice QA results for
Llama-3.1-8B-Instruct under GA. Accuracy is evaluated on the base model
and the corresponding already-trained unlearned checkpoints. The
accuracy decrease follows the same L1--L3 ordering as the PPL ratios.
}
\label{tab:qa_probe}
\end{table}

\section{\ref{rq2} Pre-unlearn Impact Auditing}
\label{sec:audit}

The previous section shows that three-layer unlearning impact varies
substantially with the choice of forget data, making forget-set selection a critical
design decision. However, this impact is only observable after running
unlearning, which is inefficient when multiple candidate forget sets must be
tested to achieve a desirable forget-retain trade-off. We therefore ask whether
we can audit, \textit{before unlearning}, which forget--evaluation pairs are likely to experience larger post-unlearning degradation. Moreover, this audit is not intended to replace final evaluation, but to provide an early sanity check and also to identify predictive features that are consistently associated with future collateral damage, providing deeper understanding on the collateral damage explored in \ref{rq1}.



\subsection{Problem Formulation}

\label{sec:audit-problem}


We formulate pre-unlearning auditing as a supervised prediction problem over
forget--evaluation pairs. Each audit example is a pair $(D_f,D_e)$, where
$D_f$ is the candidate forget set and $D_e$ is an evaluation set whose future
degradation we want to estimate before unlearning. We constrain the auditor to have no access to target LLM's gradients, unlearned checkpoints, or any post-unlearning measurements as input. For a fixed target model $M_\theta$ and unlearning algorithm $\mathcal{U}$, we extract pre-unlearning
features:
\[
\mathbf{x}_{f,e}
=
\phi(D_f,D_e,M_\theta),
\]
describing the forget set, the evaluation set, their interaction. The auditor then learns a predictor that predict an estimated \textit{collateral damage ratio}:
\[
\hat{\rho}_{f,e}
=
h_\psi(\mathbf{x}_{f,e}),
\]
which is trained by optimize a MSE loss:
\[
\min_{\psi}
\sum_{(D_f,D_e)}
\left(
h_\psi(\mathbf{x}_{f,e})-\rho_{f,e}
\right)^2 .
\]
This pair-level formulation lets the auditor ask whether a forget set is risky
in isolation, and which evaluation sets are most exposed to that forget set.

We define the prediction target $\rho_{f,e}$ as a collateral damage ratio:
\begin{equation}
\rho_{f,e}
=
\frac{
\log\!\left(\PPL^{\mathrm{after}}_{e}/\PPL^{\mathrm{before}}_{e}\right)
}{
\log\!\left(\PPL^{\mathrm{after}}_{f}/\PPL^{\mathrm{before}}_{f}\right)
}.
\label{eq:leakage_ratio}
\end{equation}
This ratio measures how much degradation leaks from the forget set to the
evaluation set, normalized by the achieved forgetting strength. A value
$\rho{\approx}0$ indicates selective unlearning or minimal collateral damage to $D_e$, $\rho{\approx}1$ indicates
uniform degradation, and $\rho{>}1$ flags cases where the evaluation set is harmed
more than the targeted forget set.

\subsection{Experimental Setup}

\paragraph{Training Supervised Regression Model}
\label{sec:audit-regressors}

We train three regression classifiers $h_\psi(\cdot)$: ridge regression, random forests
(RF), and gradient-boosted trees (XGBoost). All three predict the collateral damage ratio from the $80$-dimensional feature
vector described in \S\ref{sec:features}.

\setlength{\tabcolsep}{10pt}
\begin{table*}[tb]
\centering
\footnotesize
\begin{tabular}{c ll ccc ccc}
\toprule
\multirow{2}{*}{Model} & \multirow{2}{*}{Algorithm} & \multirow{2}{*}{Regressor}
& \multicolumn{3}{c}{LODO CV} & \multicolumn{3}{c}{Held-out Test} \\
\cmidrule(lr){4-6}\cmidrule(lr){7-9}
& & & RMSE$\downarrow$ & MAE$\downarrow$ & $R^{2}\uparrow$
& RMSE$\downarrow$ & MAE$\downarrow$ & $R^{2}\uparrow$ \\
\midrule

\multirow{9}{*}{\rotatebox{90}{Llama-3.1-8B}}
& \multirow{3}{*}{GA}
& Ridge   & 0.0725 & 0.0572 & 0.7487 & 0.0825 & 0.0651 & 0.6439 \\
& & RF      & \cellcolor{bestcell}\textbf{0.0621} & \cellcolor{bestcell}\textbf{0.0497} & \cellcolor{bestcell}\textbf{0.8153} & 0.0911 & 0.0740 & 0.5663 \\
& & XGB     & 0.0661 & 0.0521 & 0.7907 & \cellcolor{bestcell}\textbf{0.0749} & \cellcolor{bestcell}\textbf{0.0621} & \cellcolor{bestcell}\textbf{0.7067} \\
\cmidrule(lr){2-9}

& \multirow{3}{*}{NPO}
& Ridge   & 0.0751 & 0.0500 & 0.5464 & 0.0867 & \cellcolor{bestcell}\textbf{0.0645} & 0.5420 \\
& & RF      & 0.0486 & 0.0380 & 0.8101 & \cellcolor{bestcell}\textbf{0.0833} & 0.0649 & \cellcolor{bestcell}\textbf{0.5775} \\
& & XGB     & \cellcolor{bestcell}\textbf{0.0473} & \cellcolor{bestcell}\textbf{0.0368} & \cellcolor{bestcell}\textbf{0.8202} & 0.0850 & 0.0661 & 0.5599 \\
\cmidrule(lr){2-9}

& \multirow{3}{*}{UNDIAL}
& Ridge   & 0.0665 & 0.0511 & 0.7584 & 0.0984 & 0.0843 & 0.6058 \\
& & RF      & 0.0527 & 0.0398 & 0.8482 & \cellcolor{bestcell}\textbf{0.0567} & \cellcolor{bestcell}\textbf{0.0464} & \cellcolor{bestcell}\textbf{0.8691} \\
& & XGB     & \cellcolor{bestcell}\textbf{0.0523} & \cellcolor{bestcell}\textbf{0.0391} & \cellcolor{bestcell}\textbf{0.8503} & 0.0665 & 0.0570 & 0.8199 \\

\midrule

\multirow{9}{*}{\rotatebox{90}{Qwen2.5-7B}}
& \multirow{3}{*}{GA}
& Ridge   & 0.0549 & 0.0409 & 0.7948 & \cellcolor{bestcell}\textbf{0.0683} & \cellcolor{bestcell}\textbf{0.0591} & \cellcolor{bestcell}\textbf{0.7488} \\
& & RF      & \cellcolor{bestcell}\textbf{0.0431} & \cellcolor{bestcell}\textbf{0.0322} & \cellcolor{bestcell}\textbf{0.8738} & 0.0965 & 0.0865 & 0.4993 \\
& & XGB     & 0.0432 & 0.0327 & 0.8733 & 0.1173 & 0.1073 & 0.2599 \\
\cmidrule(lr){2-9}

& \multirow{3}{*}{NPO}
& Ridge   & 0.0967 & 0.0665 & 0.3397 & 0.0855 & 0.0648 & 0.3913 \\
& & RF      & \cellcolor{bestcell}\textbf{0.0813} & \cellcolor{bestcell}\textbf{0.0634} & \cellcolor{bestcell}\textbf{0.5335} & 0.0765 & 0.0594 & 0.5133 \\
& & XGB     & 0.0854 & 0.0648 & 0.4855 & \cellcolor{bestcell}\textbf{0.0614} & \cellcolor{bestcell}\textbf{0.0494} & \cellcolor{bestcell}\textbf{0.6863} \\
\cmidrule(lr){2-9}

& \multirow{3}{*}{UNDIAL}
& Ridge   & 0.1239 & 0.0897 & 0.6234 & 0.0762 & 0.0606 & 0.8579 \\
& & RF      & 0.0853 & 0.0602 & 0.8217 & \cellcolor{bestcell}\textbf{0.0677} & \cellcolor{bestcell}\textbf{0.0478} & \cellcolor{bestcell}\textbf{0.8877} \\
& & XGB     & \cellcolor{bestcell}\textbf{0.0825} & \cellcolor{bestcell}\textbf{0.0587} & \cellcolor{bestcell}\textbf{0.8331} & 0.0715 & 0.0518 & 0.8749 \\
\bottomrule
\end{tabular}
\caption{Audit-model performance excluding Spearman correlation. Best score within each model, algorithm, and split block is highlighted and shown in bold.}
\label{tab:audit-performance-main}
\end{table*}

\paragraph{Evaluation Protocol.} 
To assess whether the trained audit regressor $h_\psi(\cdot)$ generalizes beyond forget sets seen during training, we use two complementary splits. First, LODO CV performs leave-one-domain-out evaluation over the $K$ domain clusters obtained from HDBSCAN \cite{hdbscan} across the 90 training \texttt{forget\_set} groups of 9 domains. For each held-out domain $G_i$, we train $h_\psi(\cdot)$ on all forget sets drawn from the remaining $K{-}1$ domains and evaluate it on every forget set belonging to $G_i$. This ensures that no forget set in the test split shares the same domain with any training example, eliminating domain-level leakage. Reported LODO results aggregate predictions across all $K$ folds, so every forget set is evaluated exactly once under a model that has never seen its domain. Second, Held-out Test evaluates generalization to a fully held-out semantic domain.

\subsection{Feature Engineering}
\label{sec:features}

Each row of our auditing dataset corresponds to a forget set and eval set pair. We engineer features of each pair organized into the three
families summarized in Table~\ref{tab:feature-families} in Appendix: intrinsic
features of $D_f$ and $D_e$, pair-interaction features
that capture relationships between the two sets. The first three families are entirely model-agnostic, depending only on the raw texts and their
sentence-transformer embeddings.
The two set-intrinsic features
share identical definitions, applied independently to $D_f$ and $D_e$.
Each features covers five complementary aspects of the text  describing how
large, how diverse, how tight, and how low-dimensional the set is. The
pair features then exposes quantities that cannot be
recovered from $D_f$ and $D_e$ in isolation:
three features describe the centroid relation between $D_f$ and $D_e$
(cosine similarity, Euclidean distance, and norm ratio), with the cosine
term serving as a direct proxy for topical overlap, and the remaining features
are cross-set $e/f$ ratios of representative scalar descriptors.

\subsection{Results}
\label{sec:audit-results}

\paragraph{Overall Auditability.} Table~\ref{tab:audit-performance-main} shows that auditability depends on both the unlearning algorithm and the base model. Across models, UNDIAL is the easiest objective to predict, NPO the hardest, and GA intermediate, but the best regressor family and the held-out degradation pattern vary by model.

UNDIAL is the most stable case: held-out predictability remains high on both
models and is relatively insensitive to regressor choice, suggesting that its
damage aligns well with our geometry and length features. GA shows the strongest
algorithm--model interaction. On Llama, tree-based models preserve their LODO advantage on held-out domains, whereas on Qwen, ridge generalizes most stably:
rank ordering remains reliable, but error magnitude drifts across domains. NPO
is the most difficult and regressor-sensitive objective on both models, with
different failure modes across Llama and Qwen.

Overall, the LODO--held-out gap widens from UNDIAL to GA to NPO, but performance
does not collapse to chance. This indicates that the auditor captures
cross-domain structure rather than memorizing domain identity, supporting our
view that collateral damage depends on dataset geometry.

\paragraph{Feature Importance Analysis.}

\setlength{\tabcolsep}{3pt}
\begin{table}[tb!]
\centering
\footnotesize
\begin{tabular}{ccccc}
\toprule
Family & Feature & coef & z & P$>|z|$ \\
\midrule
Pair & Centroid Euclidean dist. & -0.077 & -33.014 & 0.000 \\
 & Centroid cosine sim. & 0.016 & 7.440 & 0.000 \\
 & Mean word count ratio & -0.037 & -5.887 & 0.000 \\
 & Top-1 variance ratio & 0.021 & 5.809 & 0.000 \\
 & Type-token ratio & -0.026 & -4.520 & 0.000 \\
\midrule
Forget & Char length & 0.044 & 20.059 & 0.000 \\
 & Word count & -0.050 & -12.849 & 0.000 \\
 & Char length & 0.044 & 10.960 & 0.000 \\
 & Pairwise emb.\ sim.\ & 0.061 & 10.578 & 0.000 \\
 & Char length  & 0.015 & 8.265 & 0.000 \\
\midrule
Eval & Pairwise emb.\ sim.\ & -0.036 & -7.347 & 0.000 \\
 & Char length & 0.011 & 5.832 & 0.000 \\
 & Word count & -0.020 & -5.185 & 0.000 \\
 & Nearest-neighbor sim.\ & -0.022 & -5.176 & 0.000 \\
 & Spread / centroid norm & 0.025 & 4.465 & 0.000 \\
\bottomrule
\end{tabular}
\caption{Top-5 most predictive features per category, including \textbf{Pair} (cross-set forget--evaluation geometry), \textbf{Forget} (intra-forget statistics), and \textbf{Eval} (intra-evaluation statistics), in the audit predictor's ridge regression.  All coefficients are statistically significant. The sign indicates direction of effect on predicted collateral damage, and $|z|$ reflects standardized importance.}
\label{tab:audit-predictor-coefficients}
\end{table}

Table~\ref{tab:audit-predictor-coefficients} reports regression coefficients for the most predictive features in our collateral-damage audit predictor, grouped into three families: Pair features, Forget features, and Eval features. All listed coefficients are highly significant. The dominant signal comes from Pair features, where the centroid Euclidean distance between forget and eval sets carries by far the largest effect: the farther apart the two sets sit in representation space, the smaller the collateral damage. This is reinforced by the positive coefficient on centroid cosine similarity and the negative coefficients on cross-set ratio features , indicating that surface- and geometry-level proximity between forget and eval jointly drives utility loss. Within the Forget family, longer documents and heavier similarity tails increase damage, suggesting that long, homogeneous forget clusters exert broader influence on the model. Eval-only features contribute smaller effects, suggesting that the relative geometry between forget and eval, rather than either set's intrinsic structure, is the primary determinant of collateral damage, in line with the fixed-evaluation-manifold view underlying our framework.

\paragraph{Centroid-only control.}
Because centroid Euclidean distance is the strongest individual feature
in the full auditor, we next test whether centroid-level geometry alone
accounts for its predictive performance. We conduct this focused control
experiment on Llama-3.1-8B-Instruct. The centroid-only control uses the
same prediction target, leave-one-domain-out evaluation protocol, and
regressors as the full auditor. As shown in Table~\ref{tab:centroid_control}, centroid-level geometry alone provides
substantial predictive signal. However, the full feature set yields
substantially higher correlations for GA and NPO, while the two auditors
perform comparably for UNDIAL. These results suggest that centroid
geometry is informative but does not fully account for the auditor's
predictive performance in the Llama-3.1-8B-Instruct setting.

\begin{table}[t]
\centering
\small
\begin{tabular}{@{}lrr@{}}
\toprule
Unlearning method & Centroid-only & Full auditor \\
\midrule
GA     & 0.426 & 0.541 \\
NPO    & 0.353 & 0.557 \\
UNDIAL & 0.597 & 0.591 \\
\bottomrule
\end{tabular}
\caption{
Centroid-only control results for Llama-3.1-8B-Instruct under the same
leave-one-domain-out evaluation protocol as the full auditor. Values are
Spearman correlations between predicted and observed collateral-damage
targets; higher is better.
}
\label{tab:centroid_control}
\end{table}

\paragraph{Audit Sensitivity Analysis.}


\begin{figure}[tb!]
    \centering
    \includegraphics[width=\linewidth]{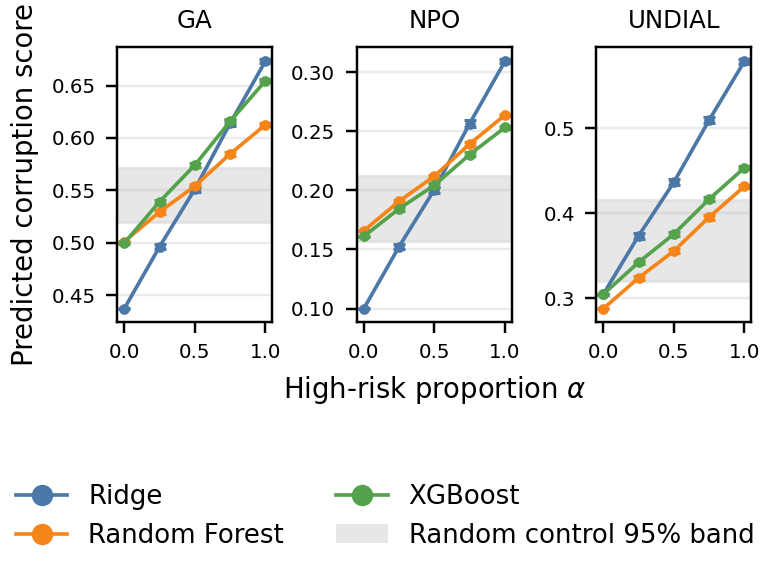}
    \caption{
    Sensitivity of fixed corruption regressors to evaluation composition.
    Each panel varies the fraction $\alpha$ of high-risk evaluation examples.
    Error bars show 95\% bootstrap confidence intervals over sampling seeds.
    The gray band shows the 95\% range induced by random resampling of the
    held-out evaluation set without controlling the high-risk proportion.
    }
    \label{fig:sensitivity-analysis}
\end{figure}

To test whether the auditor responds to meaningful changes in the evaluation set rather than random sampling noise, we keep each trained regressor fixed and vary only the risk composition of the held-out evaluation set. We construct mixtures with different high-risk proportions $\alpha$, where $\alpha{=}0$ contains only low-risk blocks and $\alpha{=}1$ contains only high-risk blocks, while keeping the number of evaluation blocks per forget set fixed. As a control, we also randomly resample evaluation blocks without controlling the high-risk proportion. If the auditor is sensitive to evaluation risk, its predicted corruption score should increase as $\alpha$ grows. Figure~\ref{fig:sensitivity-analysis} shows exactly this pattern: across GA, NPO, and UNDIAL, all regressors produce steadily increasing scores, while random resampling stays within the gray control band. This suggests that the auditors capture meaningful shifts in evaluation-set
vulnerability rather than random variation.

\section{Potential Application}

\begin{figure}[tb!]
    \centering
    \includegraphics[width=\linewidth]{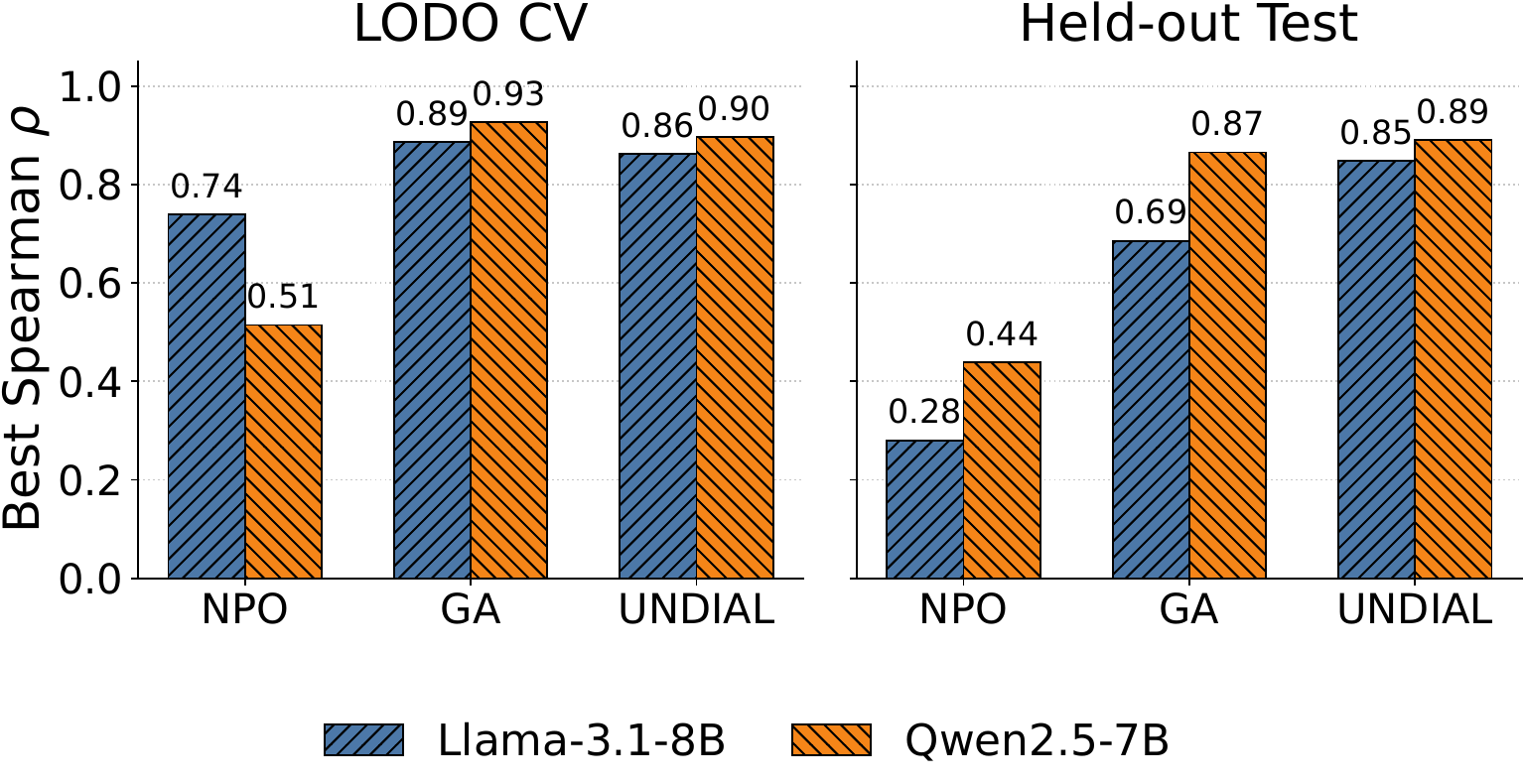}
    \caption{Best Spearman correlation across the three regressors for each model, algorithm, and evaluation settings.}
    \label{fig:audit-spearman-best-bars}
\end{figure}

In practice, the auditor is most useful not as an exact simulator of post-unlearning perplexity, but as a pre-unlearning ranking tool. Given multiple candidate forget--evaluation pairs, practitioners typically face three triage questions: which pairs should be inspected first, which forget sets are likely to cause broader collateral damage, and where limited evaluation or retain-data budgets should be allocated. Figure~\ref{fig:audit-spearman-best-bars} reports Spearman rank correlation between predicted and observed collateral damage across the three regressors. Under LODO cross-validation, the best auditor reaches $\rho=0.93$ on GA and $\rho=0.86$ to $0.90$ on UNDIAL across both base models, and stays at $\rho \geq 0.51$ even on the harder NPO objective. On the fully held-out domain, ranking quality drops as expected but remains informative for UNDIAL and for GA, where Qwen generalizes notably better than Llama, while NPO degrades more sharply. This pattern indicates that the auditor captures cross-domain risk structure rather than memorizing domain identity, and that the ordering signal is most reliable precisely for the algorithms whose collateral damage is most tightly coupled to forget--evaluation geometry. 

Even when exact magnitude prediction varies across domains, a reliable ordering is sufficient for triage: users can prioritize the top-ranked risky pairs for full unlearning evaluation, add retain data around vulnerable domains, or reject risky forget-set choices early in the pipeline before any optimization is run. The application of pre-unlearning auditing is therefore to turn post-hoc damage measurement, which requires one unlearning run per candidate forget set, into a cheaper ranking problem whose marginal cost is dominated by feature extraction, allowing large-scale unlearning pipelines to be guided by upfront risk estimates rather than purely post-hoc evaluation.

\section{Conclusion}
\label{sec:conclusion}
We formalize unlearning impact as a three-layer profile and show that damage is not confined to the intended target: across WikiText forget sets, it decays with distance from the forget data but does not disappear, and varies substantially under fixed hyper-parameters, making the forget set itself an object of audit rather than a static benchmark. Asking what can be known before unlearning, we find the strongest audit signals are not intrinsic properties of the forget set, but cross-set features comparing forget and evaluation data (centroid similarity, centroid distance, lexical and length ratios), and their importance is stable across unlearning algorithms, suggesting collateral damage is partly determined by pre-existing coupling between the two. We therefore position forget-set auditing as a screening tool that surfaces interpretable risk factors and prioritizes which candidates deserve expensive audit-then-unlearn, rather than replacing full evaluation.


\section*{Limitations}
\label{sec:limitations}

\textbf{Model and algorithm coverage.} Our study covers several representative unlearning objectives, including GA, NPO, and UNDIAL, but it does not exhaust the full design space of unlearning methods. The behavior of the audit features may vary across base models, model scales, instruction-tuned checkpoints, and representation-level approaches such as RMU. This is expected, since different algorithms intervene at different stages of the optimization or representation pipeline. As a result, the same forget--evaluation geometry may not induce identical collateral-damage patterns across all settings. We therefore view our results as an initial characterization of pre-unlearning auditability, rather than a universal claim about all unlearning algorithms.

\textbf{Evaluation metrics.} We measure unlearning impact mainly
through changes in perplexity. Perplexity is scalable and fine-grained, but it
is still a proxy for behavioral degradation. A model may show higher perplexity
without a large drop in downstream task performance, or may preserve perplexity
while changing answer correctness, refusal behavior, calibration, or factual
consistency. Future work should extend the audit to task-level, generation-level,
and human-judged outcomes.

\textbf{Fixed experimental design.} We hold several design choices fixed
to isolate the effect of the forget set, including hyperparameters, evaluation
construction, and the three-layer view of L1 intended, L2 same-domain, and L3
distant-domain impact. Real unlearning pipelines may tune hyperparameters per
request or face overlapping domains that do not fit cleanly into discrete layers. 


\section*{Acknowledgments}
This work used Jetstream2 at Indiana University through allocations \#CIS251092, \#CIS260332, \#CIS250090, \#CIS240570 from the Advanced Cyberinfrastructure Coordination Ecosystem: Services \& Support (ACCESS) program, which is supported by National Science Foundation grants \#2138259, \#2138286, \#2138307, \#2137603, and \#2138296. The authors thank the reviewers for their detailed feedback on this work. The authors acknowledge the use of AI for editorial assistance.

\bibliography{custom}

\appendix

\clearpage
\section{Reproducibility Details}
\label{app:repro}

\paragraph{Data.} 

To utilize WikiText-103 for our work, we first segment the corpus into passage-level units and remove passages that are too short or otherwise unsuitable for stable PPL evaluation. We then embed each passage with \texttt{all-MiniLM-L6-v2}
\citep{wang2020minilmv2}, obtaining a $384$-dimensional semantic vector per
passage. We cluster these passage embeddings to obtain coherent topical pools from which
forget sets can be sampled. Specifically, we use HDBSCAN \citep{hdbscan} with 
\texttt{min\_cluster\_size}{=}$200$, \texttt{min\_samples}{=}$5$, Euclidean distance, 
and default Excess-of-Mass (EOM) cluster selection. To This produces $10$ non-noise semantic clusters and a noise pool. We discard the noise pool and use only
the clustered passages when constructing forget-set candidates. Fig.~\ref{fig:dataset-pipeline} summarize the construction.

\begin{table}
\begin{center}
\footnotesize
\begin{tabular}{@{}ll@{}}
    \toprule
    Component & Setting \\
    \midrule
    HDBSCAN & min cluster $200$; min samples $5$ \\
    Distance & Euclidean; EOM selection \\
    Cluster sizes & $202 / 346 / 3{,}276$ (min/median/max) \\
    Triplets & $100$ triplets, seed $42$ \\
    Split size & $50$ texts each; disjoint within triplet \\
    Eligibility & cluster size at least $150$ \\
    \bottomrule
\end{tabular}
\caption{HDBSCAN clustering and triplet construction settings. Each triplet consists of mutually disjoint forget, retain, and evaluation splits, with 50 texts per split.}
\end{center}
\end{table}

\paragraph{Hyperparameters}
Hyperparameters follow defaults from OpenUnlearning
\citep{dorna2025openunlearning}: learning rate $1\!\times\!10^{-5}$ with
linear decay, paged AdamW (32-bit), per-device batch size $1$ with gradient
accumulation $4$ (effective batch $4$), $5$ epochs, weight decay $0.01$, BF16.
All runs use a single H100 80\,GB and produce $\approx 1.5$\,TB of checkpoints
($\approx 15$\,GB each) for one combination (e.g. GA+Llama-3.1-8B).

\paragraph{Evaluation.} PPL is computed in BF16 with stride $512$ and context
$1024$; each text is scored independently. Base-model PPL is cached once per
evaluation, then reused across the $100$ unlearned
checkpoints.

\paragraph{Hardware.} All runs use a single H100 (80\,GB) node.

\begin{table}[tb!]
\footnotesize
\centering
\setlength{\tabcolsep}{4pt}
\begin{tabular}{@{}p{0.18\linewidth}p{0.34\linewidth}p{0.38\linewidth}@{}}
    \toprule
    \textbf{Prefix} & \textbf{Feature Family} & \textbf{Examples} \\
    \midrule
    \texttt{f\_}
        & $D_f$ intrinsic
        & length, embedding spread \\
    \texttt{e\_}
        & $D_e$ intrinsic
        & length, lexical diversity \\
    \texttt{pair\_}
        & $D_f$--$D_e$ interaction
        & centroid distance, ratio features \\
    \bottomrule
\end{tabular}
\caption{Features Family used for pre-unlearning auditing.}
\label{tab:feature-families}
\end{table}

\section{Feature Engineering Hyper-parameters}

Hyperparameters are tuned with RandomizedSearchCV
($40$ iterations, $5$-fold GroupKFold on forget\_set,
$R^{2}$ as the inner scoring criterion). To respect the LODO protocol of
\S\ref{sec:audit-problem}, the search is run only on the $K{-}1$
training domains of each outer fold; the held-out domain is never seen
at tuning time. We report RMSE, MAE, coefficient of determination
$R^{2}$, and Spearman rank correlation $\rho$. Two evaluation splits are
distinguished throughout: LODO CV, leave-one-domain-out across
the $90$ training forget\_set cluster, and Held-out
Test, a fully held-out family of $10$ clusters whose domain never appears in any training fold.

\setlength{\tabcolsep}{1pt}
\begin{table}
\centering
\footnotesize
\begin{tabular}{llrrrr}
\toprule
Family & Feature & coef & std err & z & P$>|z|$ \\
\midrule
Pair & centroid eucl & -0.077 & 0.002 & -33.014 & 0.000 \\
 & centroid cos sim & 0.016 & 0.002 & 7.440 & 0.000 \\
 & word count mean ratio & -0.037 & 0.006 & -5.887 & 0.000 \\
 & top1 pc ratio ratio & 0.021 & 0.004 & 5.809 & 0.000 \\
 & type token ratio ratio & -0.026 & 0.006 & -4.520 & 0.000 \\
\midrule
Forget & char len p50 & 0.044 & 0.002 & 20.059 & 0.000 \\
 & word count std & -0.050 & 0.004 & -12.849 & 0.000 \\
 & char len std & 0.044 & 0.004 & 10.960 & 0.000 \\
 & pairwise sim q90 & 0.061 & 0.006 & 10.578 & 0.000 \\
 & char len p95 & 0.015 & 0.002 & 8.265 & 0.000 \\
\midrule
Eval & pairwise sim std & -0.036 & 0.005 & -7.347 & 0.000 \\
 & char len p95 & 0.011 & 0.002 & 5.832 & 0.000 \\
 & word count std & -0.020 & 0.004 & -5.185 & 0.000 \\
 & nn sim mean & -0.022 & 0.004 & -5.176 & 0.000 \\
 & spread over centroid & 0.025 & 0.006 & 4.465 & 0.000 \\
\bottomrule
\end{tabular}
\caption{Top-5 most predictive features per family in the audit predictor's ridge regression. Features are grouped into three families: \textbf{Pair} (cross-set forget--evaluation geometry), \textbf{Forget} (intra-forget statistics), and \textbf{Eval} (intra-evaluation statistics). All listed coefficients are statistically significant ($p < 0.001$). The sign indicates direction of effect on predicted collateral damage, and $|z|$ reflects standardized importance.}
\label{tab:audit-predictor-coefficients}
\end{table}

\end{document}